\documentclass[conference]{IEEEtran}
\IEEEoverridecommandlockouts
\makeatletter
\newcommand{\linebreakand}{%
  \end{@IEEEauthorhalign}
  \hfill\mbox{}\par
  \mbox{}\hfill\begin{@IEEEauthorhalign}
}
\makeatother
\usepackage{mwe} 
\usepackage{subcaption}
\usepackage{cite}
\usepackage{amsmath,amssymb,amsfonts}
\usepackage{graphicx}
\usepackage{textcomp}
\usepackage{xcolor}
\usepackage{algpseudocode}
\usepackage{algorithm}
\def\BibTeX{{\rm B\kern-.05em{\sc i\kern-.025em b}\kern-.08em
    T\kern-.1667em\lower.7ex\hbox{E}\kern-.125emX}}
\begin{document}

\title{AssistTaxi: A Comprehensive Dataset for Taxiway Analysis and Autonomous Operations\\
}
\author{
\IEEEauthorblockN{Parth Ganeriwala}
\IEEEauthorblockA{\textit{Dept. of Computer Science} \\
\textit{Florida Institute of Technology}\\
Melbourne, Florida \\
pganeriwala2022@my.fit.edu}
\and
\IEEEauthorblockN{Siddhartha Bhattacharyya}
\IEEEauthorblockA{\textit{Dept. of Computer Science} \\
\textit{Florida Institute of Technology}\\
Melbourne, Florida \\
sbhattacharyya@fit.edu}
\and
\IEEEauthorblockN{Sean Gunther}
\IEEEauthorblockA{\textit{Dept. of Aerospace Engineering} \\
\textit{Florida Institute of Technology}\\
Melbourne, Florida \\
sgunther2018@my.fit.edu}
\linebreakand
\IEEEauthorblockN{Brian Kish}
\IEEEauthorblockA{\textit{Dept. of Mechanical Engineering} \\
\textit{Univeristy of Michigan}\\
Ann Arbor, Michigan \\
bakish@umich.edu}
\and
\IEEEauthorblockN{Mohammed Abdul Hafeez Khan}
\IEEEauthorblockA{\textit{Dept. of Computer Science} \\
\textit{BITS Pilani, Dubai Campus}\\
Dubai, United Arab Emirates \\
f20190091@dubai.bits-pilani.ac.in}
\and
\IEEEauthorblockN{Ankur Dhadoti}
\IEEEauthorblockA{\textit{Dept. of Computer Science} \\
\textit{Florida Institute of Technology}\\
Melbourne, Florida \\
adhadoti2019@my.fit.edu}
\linebreakand
 \IEEEauthorblockN{Natasha Neogi}
 \IEEEauthorblockA{\textit{Langley Research Center} \\
 \textit{NASA}\\
 Hampton, Virginia \\
 natasha.a.neogi@nasa.gov}
}


\maketitle

\begin{abstract}
The availability of high-quality datasets play a crucial role in advancing research and development especially, for safety critical and autonomous systems. In this paper, we present AssistTaxi, a comprehensive novel dataset which is a collection of images for runway and taxiway analysis. The dataset comprises of more than 300,000 frames of diverse and carefully collected data, gathered from Melbourne (MLB) and Grant-Valkaria (X59) general aviation airports. The importance of AssistTaxi lies in its potential to advance autonomous operations, enabling researchers and developers to train and evaluate algorithms for efficient and safe taxiing. 
Researchers can utilize AssistTaxi to benchmark their algorithms, assess performance, and explore novel approaches for runway and taxiway analysis. Additionally, the dataset serves as a valuable resource for validating and enhancing existing algorithms, facilitating innovation in autonomous operations for aviation. We also propose an initial approach to label the dataset using a contour based detection and line extraction technique. 
\end{abstract}


\section{Introduction}
In recent years, autonomous taxiing in aviation has emerged as a transformative technology, revolutionizing the way aircraft maneuver on taxiways and runways \cite{Nejad2021}. The convergence of advancements in robotics, computer vision, and artificial intelligence has paved the way for the development and deployment of autonomous systems in aviation. These systems leverage sophisticated algorithms and sensor technologies to navigate complex airport environments, ensuring safe and efficient taxiing operations \cite{Zhang2020}.

Traditionally, taxiing operations have relied on human pilots to navigate aircraft on the ground, requiring meticulous coordination with air traffic control and adherence to strict operational procedures. However, with the advent of autonomous taxiing, aircraft can now navigate taxiways and runways with reduced human intervention, leading to the possibility of increased operational efficiency and potentially improved safety. Autonomous taxiing systems make use of a variety of technologies, including computer vision, LiDAR, radar, and advanced navigation algorithms \cite{Zhang2018}. Computer vision algorithms analyze visual data captured by onboard cameras or external sensors to identify and track runway markings, signs, and other aircraft or objects in the vicinity \cite{alonso2023automatic}. LiDAR and radar sensors provide additional depth perception and obstacle detection capabilities, ensuring safe navigation in complex environments \cite{kong2015ground}. These autonomous systems are equipped with intelligent decision-making algorithms that can process real-time data, assess environmental conditions, and optimize taxi routes. By leveraging machine learning techniques, these algorithms can adapt and learn from previous experiences, enhancing their ability to make informed decisions in varying airport scenarios. Machine learning algorithms heavily rely on data to train models and make informed decisions \cite{emmert2022taxonomy}. By leveraging vast amounts of historical and real-time data collected from airport operations, weather conditions, aircraft movements, and communication protocols, machine learning algorithms can learn patterns, predict potential hazards, and optimize taxiing operations for enhanced safety. 

In the context of taxiway analysis and autonomous operations, the availability of a comprehensive and meticulously annotated dataset is of primary importance. Towards this end, we present AssistTaxi \cite{parth_ganeriwala_2023_8144439}, a robust dataset specifically curated to address the critical need for high-quality data in the field. AssistTaxi is designed to capture real-world scenarios and encompasses data collected from Melbourne Orlando International Airport (KMLB) and Grant-Valkaria (X59) general aviation airport. By focusing on diverse aspects of taxiway operations, including aircraft movements, and environmental factors, AssistTaxi offers researchers and developers a resource for analysis, benchmarking, and algorithm development.
The research paper  consists of Section \ref{relatedwork} that details the related work while Sections \ref{collect} and \ref{dataset} outline the data collection setup and describe our dataset. Section \ref{initial} discusses about our initial approach for labelling the images in the dataset. Lastly, Section \ref{conclusion} presents our conclusion from this research.


\section{Related Works}
\label{relatedwork}

In the past decade, the research community has been introduced with a collection of autonomous vehicular (AV) datasets, which focus on object detection, scene segmentation and navigation algorithms. Singh et al. \cite{Singh2023} presented a multi-label, multi-task dataset called ROad event Awareness Dataset (ROAD) for Autonomous Driving. Their dataset is intended to assess an autonomous vehicle's capacity to recognize road events, which are characterized as triplets consisting of an active agent, the action(s) it performs, and the scene locations. Additionally, they plan to extend their dataset's annotation methodology to other state-of-the-art datasets such as PIE \cite{rasouli2019pie} and Waymo \cite{sun2020scalability}. Shirke et al. \cite{Shirke2019} provide a comprehensive survey of the different publicly available lane datasets with a comparative study for lane detection applications. Only one of the datasets mentioned; BDD100K \cite{yu2020bdd100k}, has over 120 million images, whereas the other datasets are considerably smaller. Particularly,  datasets with a large image set prove to be useful for researchers with deep learning algorithm applications in the field of autonomous vehicular operations \cite{vargas2017deep,ganeriwalacross}.    

One of the major challenges for aircraft navigation on taxiways is line detection and tracking. Researchers \cite{Nejad2021} have tested particle filter, hough transform and neural network methodologies such as LaneNet, on both simulated images (from a product of the OKTAL-SE company) and real ones (from Airbus Operations S.A.S.). However, all these methods are affected by the lack of diversity and data in the dataset. They also provide insights into lane detection difference between the simulated and real-world images. Liu et al. \cite{liu2019vision} presented a new approach for autonomous taxiing that aimed at improving the efficiency and safety of ground operations in commercial or military airports. They formulated the research problem to be of a hybrid nature with an approach based on vision-based perception, obstacle avoidance, and feedback control theory. Still, their approach was demonstrated in a simulation with the airport environment generated, observed, and controlled using the UnrealEngine. The model knowledge has not seen real-world scenarios in its training phase, thus having a higher chance of failure in the case of unforeseen airdrome conditions. Zhang et al. \cite{Zhang2020} proposed a software architecture for autonomous taxiing operations, and tested their approach on taxiways simulated within the X-Plane flight simulator. They indicate future interest in formalizing the complete taxiing specifications into requirements represented in signal temporal logic and developing a modular formal verification framework to ensure that the proposed architecture provides system safety guarantees. This can be accomplished by employing more realistic sensor modalities, such as cameras with (possibly learning-based) perception components, which requires a vision based dataset. Knowledge enhancing framework via self-learning approaches \cite{Lu2016} have also been proposed which reinforces prior knowledge from the vision input during the taxiing process of a UAV. They have tested their framework on their created highly controlled environment but plan to test in on data from a real taxiway and runway.   

Other aviation safety applications such as airport pavement inspection \cite{alonso2023automatic}, airport runway segmentation \cite{Chen2023}, runway detection for a landing approach \cite{Amit2021, ducoffe2023lard} and foreign object debris (FOD) detection system \cite{garcia2021convolutional,Li2020} have tested their approaches on synthetic and simulated data. Boeing has been working on an experimental autonomous system for center-line tracking on airport taxiways \cite{pasareanu2023closed}. However, their dataset is proprietary with only public access to their deep neural network (DNN) system called TaxiNet. Research has been conducted in developing formal methods for system and safety analysis for TaxiNet \cite{he2021system}. Due to the lack of data, Fremont et al. \cite{fremont2020formal} used the \textit{SCENIC} probabilistic programming language, to drive tests in the X-Plane flight simulator while evaluating TaxiNet. NASA Ames Research Center presented SysAI \cite{he3system} which is flexible statistical learning framework for the V\&V and analysis of complex and high-dimensional Aerospace systems with DNN and AI components. Formal verification and validation frameworks have also been developed for the assurance of learning based aircraft taxiing \cite{cofer2020run, Cofer2020}. Torfah et al. \cite{torfah2021formal} presented \textit{VERIFAI}, an open-source toolkit for the formal design and analysis of systems that include AI/ML components. Their system was tested with perception components generated from X-Plane.  

All of the formal methods of V\&V were conducted on synthetic data generated from flight simulators such as X-Plane, Unreal Engine or proprietary data. There are no publicly available real-world datasets on taxiways and runways for autonomous aircraft operations \cite{brownlee2014airport}. Therefore, there is a need for real-world data to benchmark algorithms, models and test safety assurance frameworks.

\section{Collection Setup}
\label{collect}
\begin{figure}[h!tbp]
    \centering
    \includegraphics[width=\linewidth]{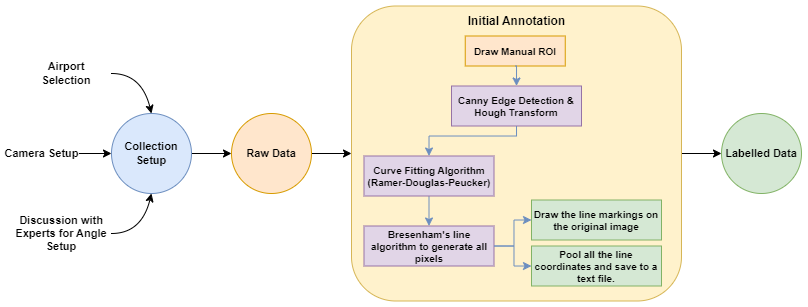}
    \caption{Dataset Collection Process Flow}
    \label{fig:dataprocess}
\end{figure}
\begin{figure}[h!tbp]
    \centering
    \includegraphics[width=0.7\linewidth]{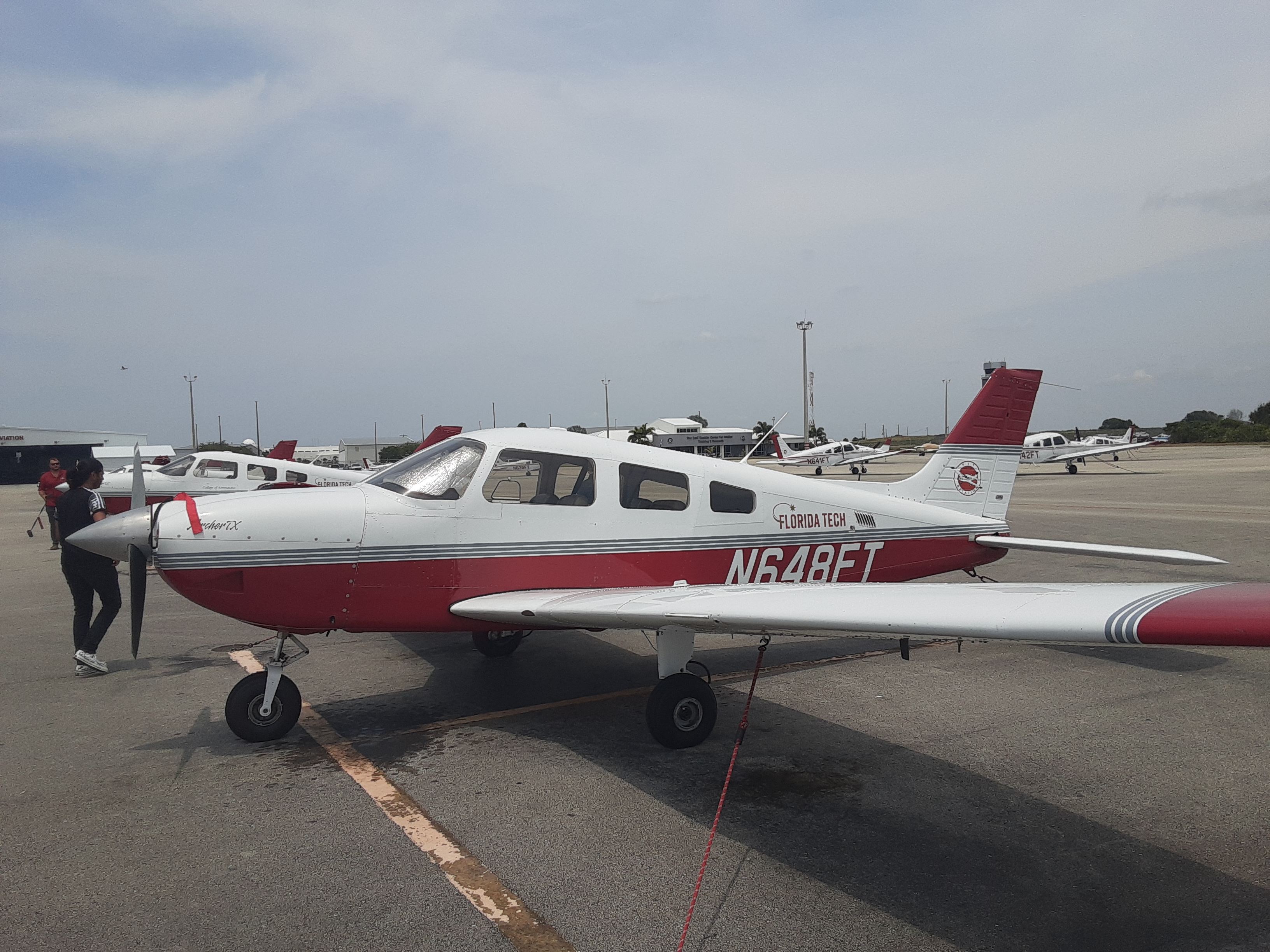}
    \caption{Piper Warrior Aircraft}
    \label{fig:aircraft}
\end{figure}
\begin{figure}[h!tbp]
\centering
  \includegraphics[width=0.8\linewidth]{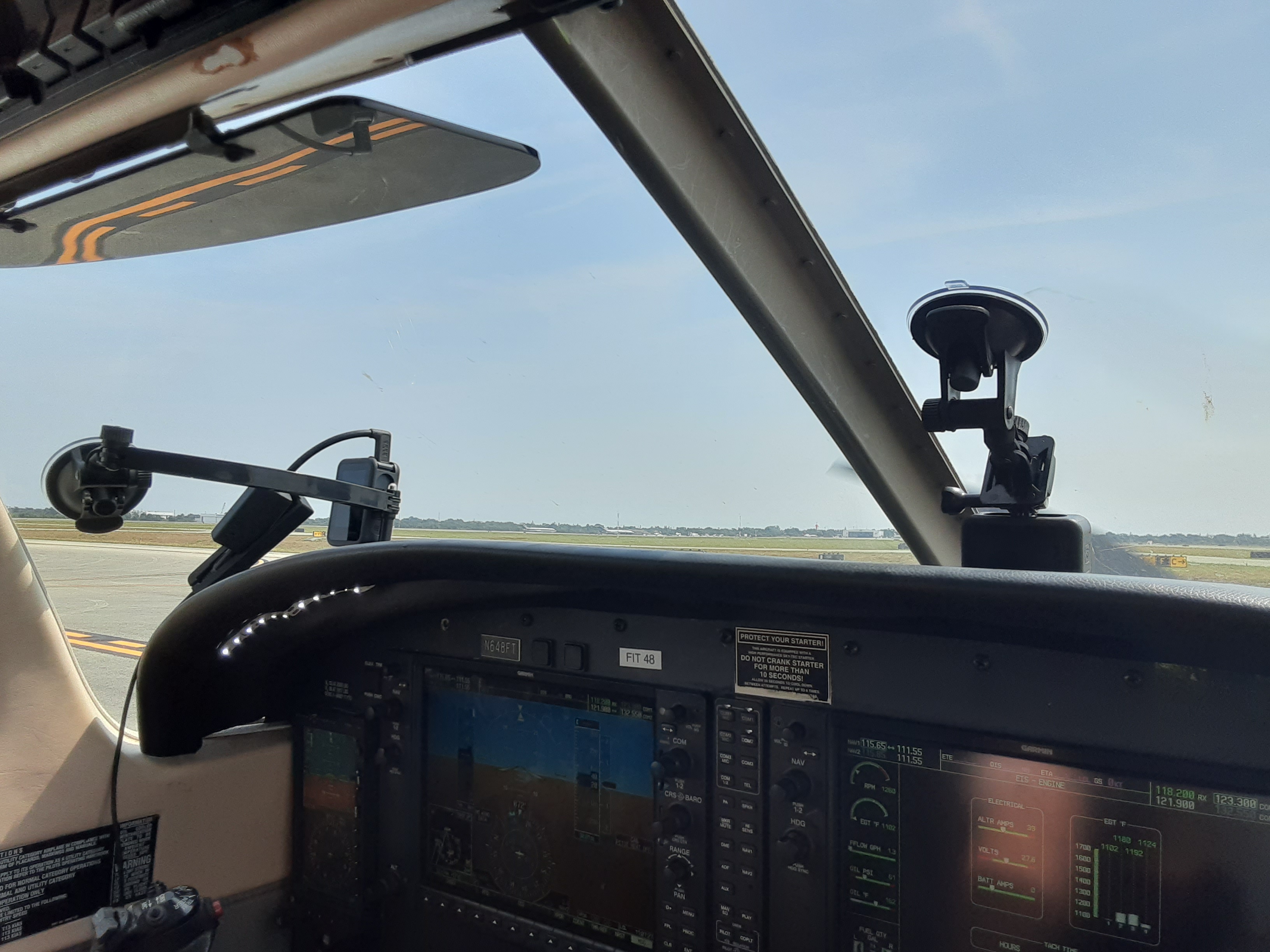}
  \caption{Setup of GoPro Cameras from the Left Side}\label{fig:left}
\end{figure}

\begin{figure}[h!tbp]
\centering
  \includegraphics[width=0.8\linewidth]{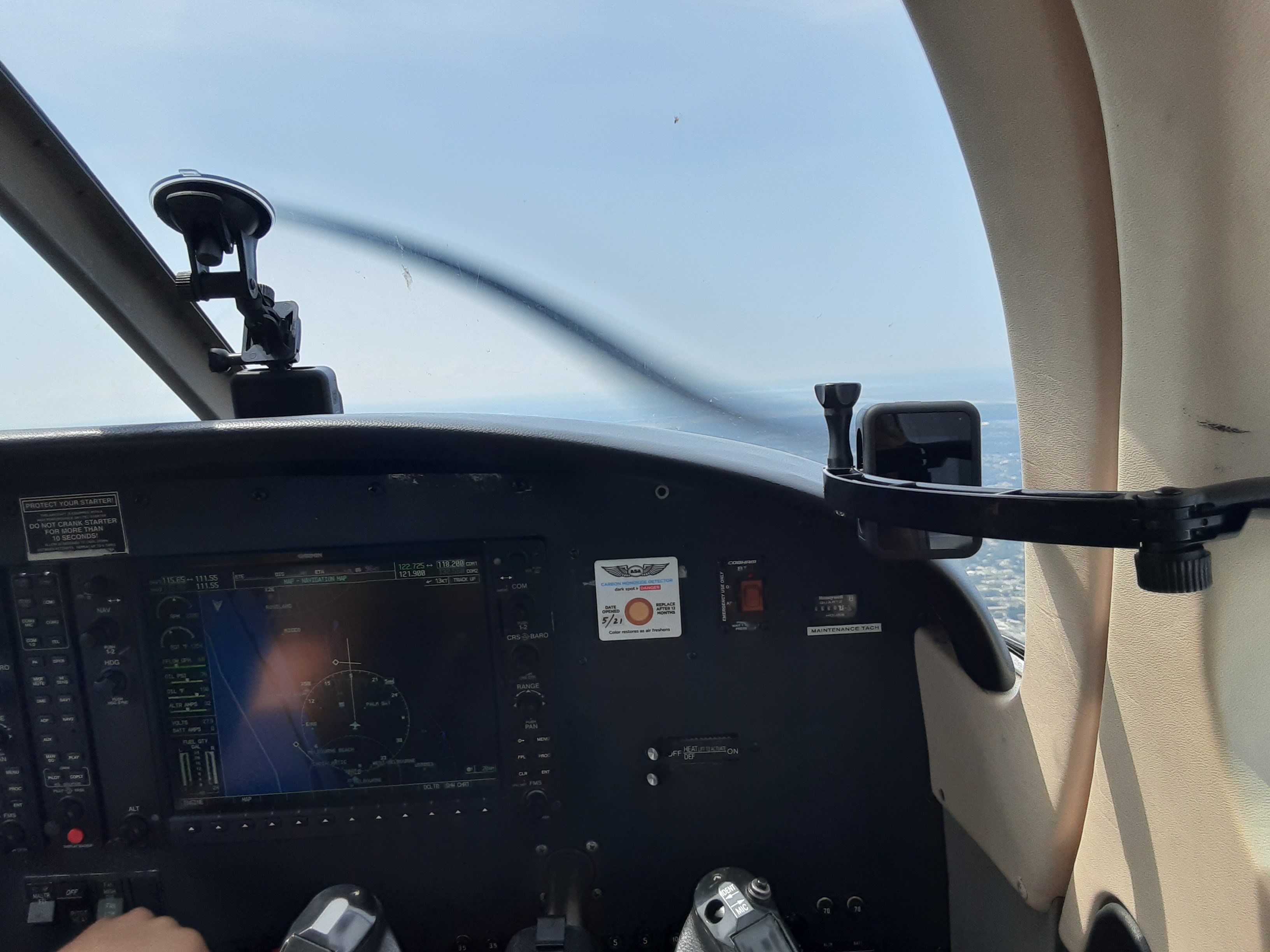}
  \caption{Setup of GoPro Cameras from the Right Side}\label{fig:right}
\end{figure}

\begin{figure}
    \centering
    \includegraphics[width=0.8\linewidth]{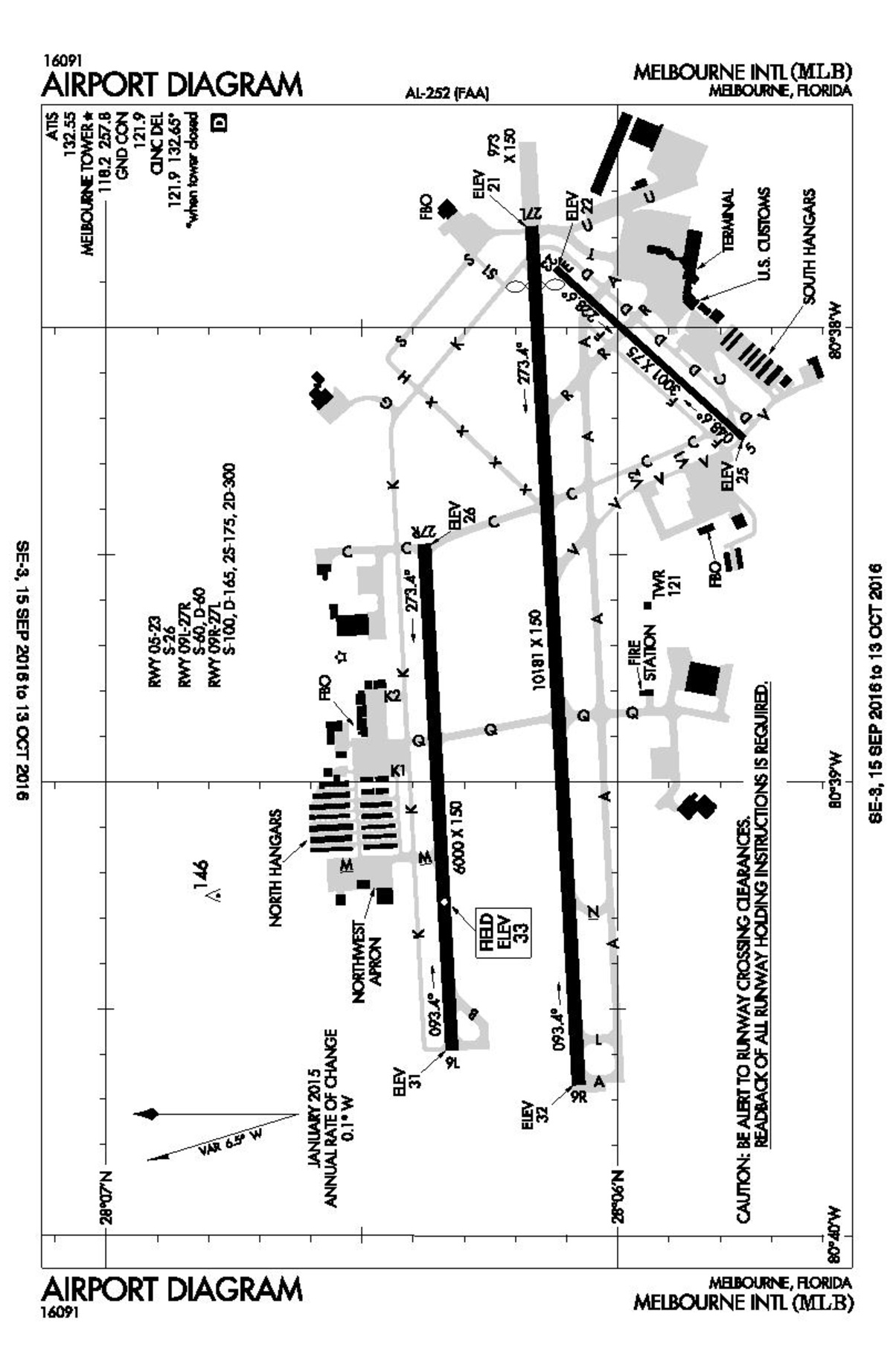}
    \caption{Airport Diagram of Melbourne Orlando International Airport}
    \label{fig:kmlb}
\end{figure}

\begin{figure}
    \centering
    \includegraphics[width=0.8\linewidth]{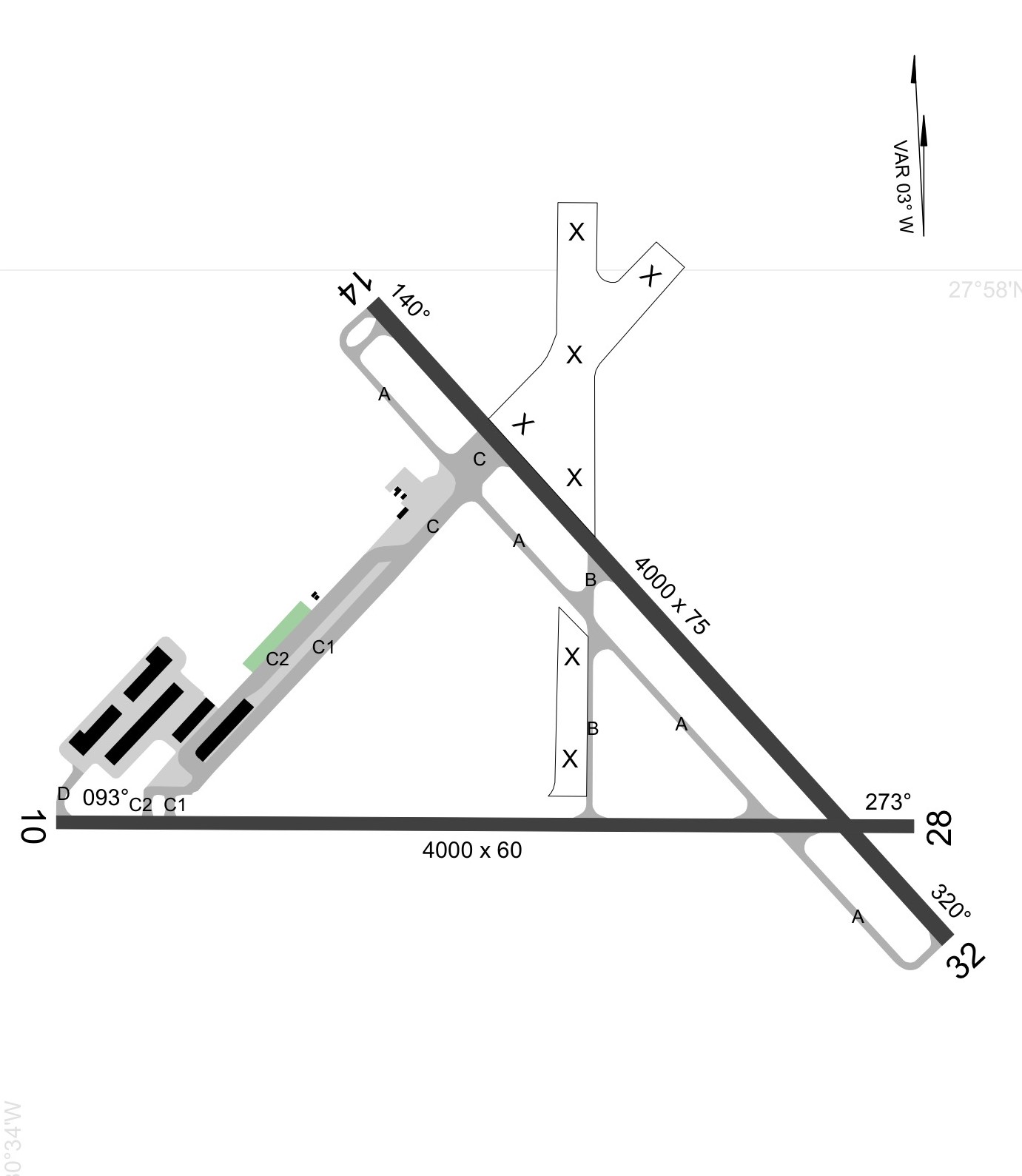}
    \caption{Airport Diagram of Valkaria Airport }
    \label{fig:x59}
\end{figure}

 
The collection process for the AssistTaxi dataset has been represented in Figure \ref{fig:dataprocess}. The AssistTaxi dataset has been recorded from a Piper Cherokee Warrior, single-engine, fixed gear aircraft as shown in Figure \ref{fig:aircraft}. Three GoPro Hero 8 cameras were fitted in the cockpit having camera angles of the center, right and left as shown in Figure \ref{fig:left} and \ref{fig:right}. This was done after discussions with aircraft experts to adjust camera angles and positions. The camera has a 12MP sensor with capabilities of recording videos in 4K/60p, 2.7K/120p and 1080/240p modes (up to 100 Mbps bit rate) utilizing its HyperSmooth 2.0 video stabilization feature. The video format can be in both H.264 and H.265 codecs. Another set of data was collected with the GoPro Hero 10 camera, with a 23MP 1/2.3-inch sensor,GP2 Processor, capable of recording 5k/60 fps, 4k/120 fps, 2.7k/240 fps videos. The data was collected from taxiing and multiple takeoff landing operations at the Melbourne Orlando International airport (KMLB) and Valkaria airports (X59). KMLB as shown in Figure \ref{fig:kmlb}, is a Class D general aviation airport, 1.5 miles northwest of downtown Melbourne in Brevard County, Florida, and 50 miles southeast of Orlando, on Florida's Space Coast. X59 as shown in Figure \ref{fig:x59}, is a Class G general aviation airport, located 1 mile west of the central business district of the city of Grant-Valkaria in Brevard County, Florida. A total of 15 videos were recorded and then frames were extracted resulting in 350,000 images. The frame extraction was carried out using a Linux system, which had the x86\_64 architecture and an Intel Core i7-9700K CPU running at 3.60GHz. The system had a total of 15GB of RAM and was running Ubuntu 22.04.1 LTS.


\section{AssistTaxi Dataset} \label{dataset}

\begin{figure}[h!tbp]
\centering
  \includegraphics[width=0.8\linewidth]{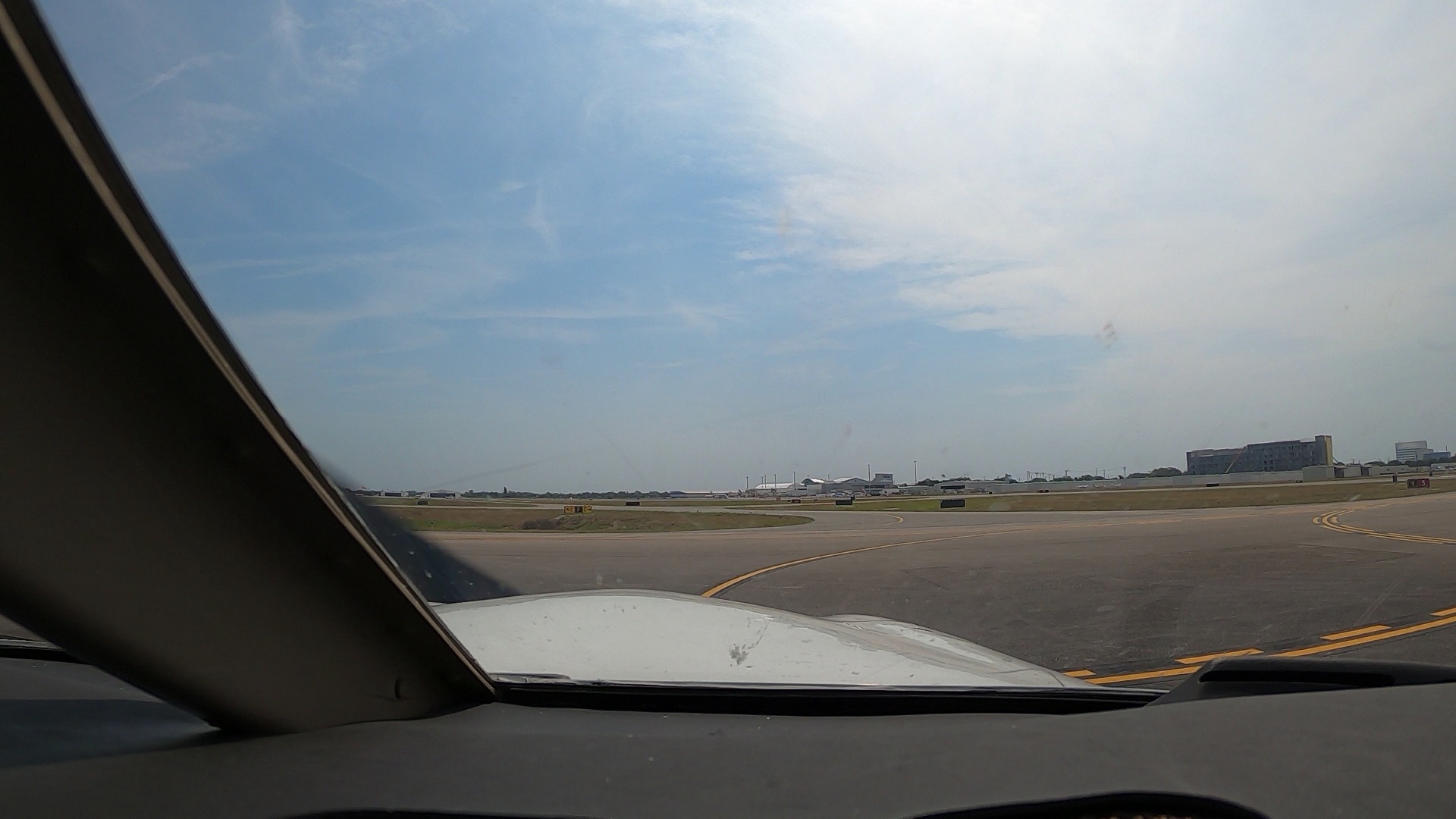}
  \caption{Taxiway Instance}\label{fig:taxiway}
\end{figure}

\begin{figure}[h!tbp]
\centering
  \includegraphics[width=0.8\linewidth]{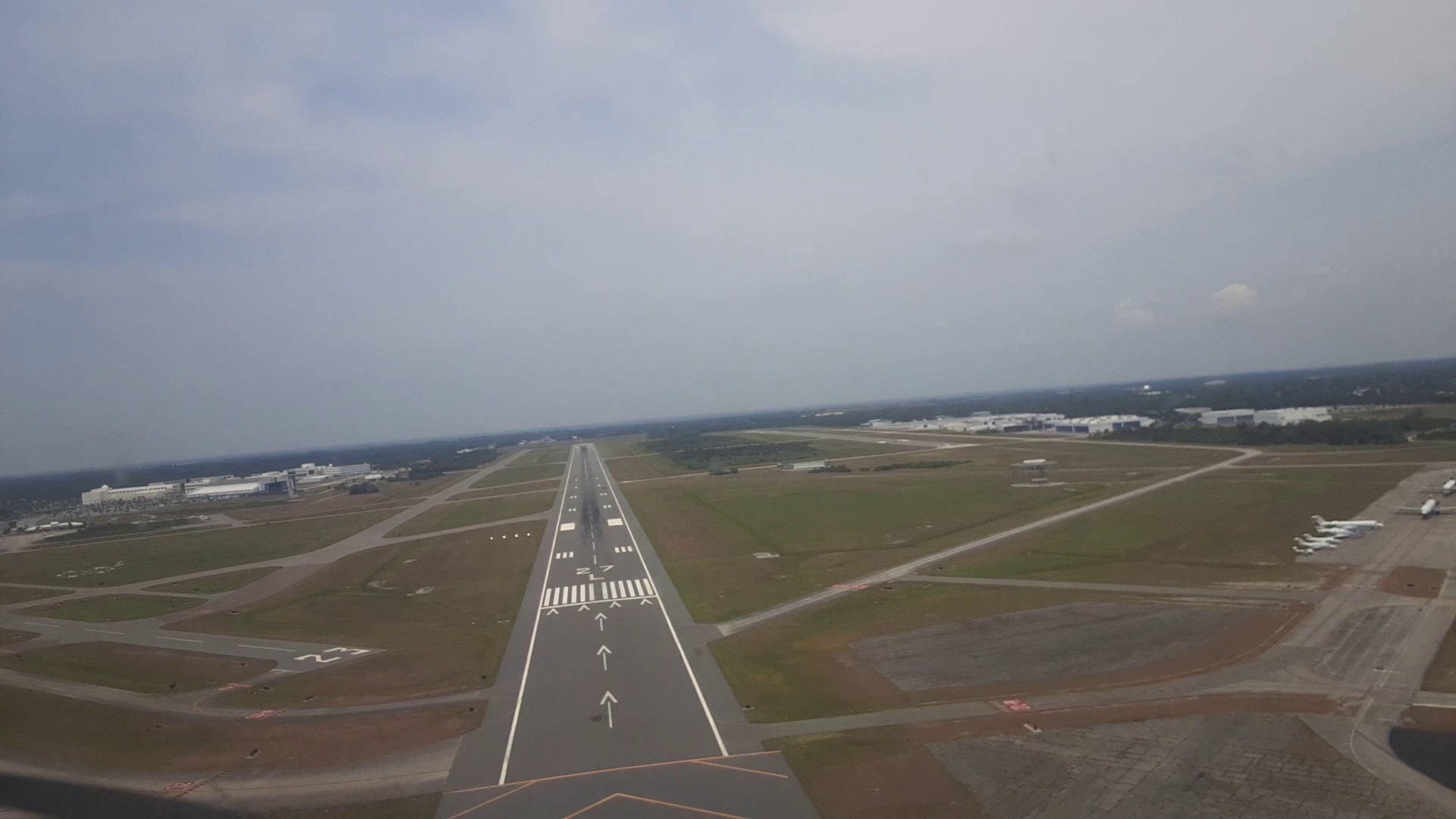}
  \caption{Runway Instance}\label{fig:runway}
\end{figure}

The raw data described in this paper can be accessed from \textit{https://github.com/ParthGaneriwala/AssistTaxi} and contains $\sim$20\% of our overall recordings. The reason for this is that primarily data with coordinate annotations has been put online, though we will make more data available upon request. Furthermore, we have removed all sequences which did not yield any annotation results. The raw data set is divided into the categories 'Taxiway' and 'Runway'. Example frames are illustrated in Figure \ref{fig:taxiway} and \ref{fig:runway}. For each sequence, we provide the raw data and coordinate annotations in form of x-y data points as collected from our initial annotation approach discussed in further sections. Our recordings have taken place on the 19th of May 2022 and on the 19th of February 2023 during daytime. The total size of the collected data is 150 GB.

\section{Initial Approach for Labeling Images: Contour-Based Detection and Line Extraction}
\label{initial}
\begin{figure*}[!htbp]
\minipage{0.32\textwidth}
  \includegraphics[width=\linewidth]{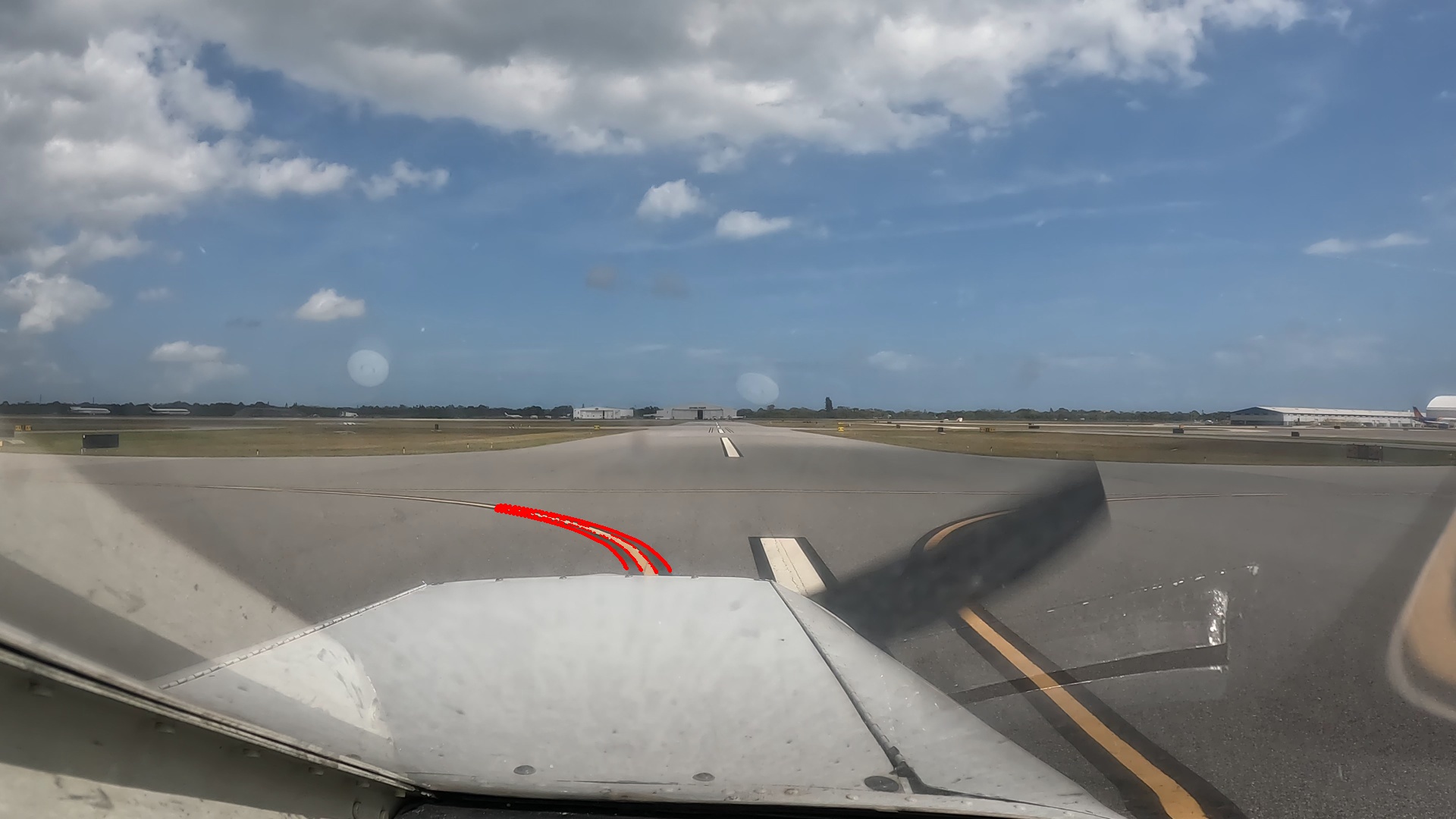}
  \caption{Left Direction}\label{fig:awesome_image2}
\endminipage\hfill
\minipage{0.32\textwidth}%
  \includegraphics[width=0.9\linewidth]{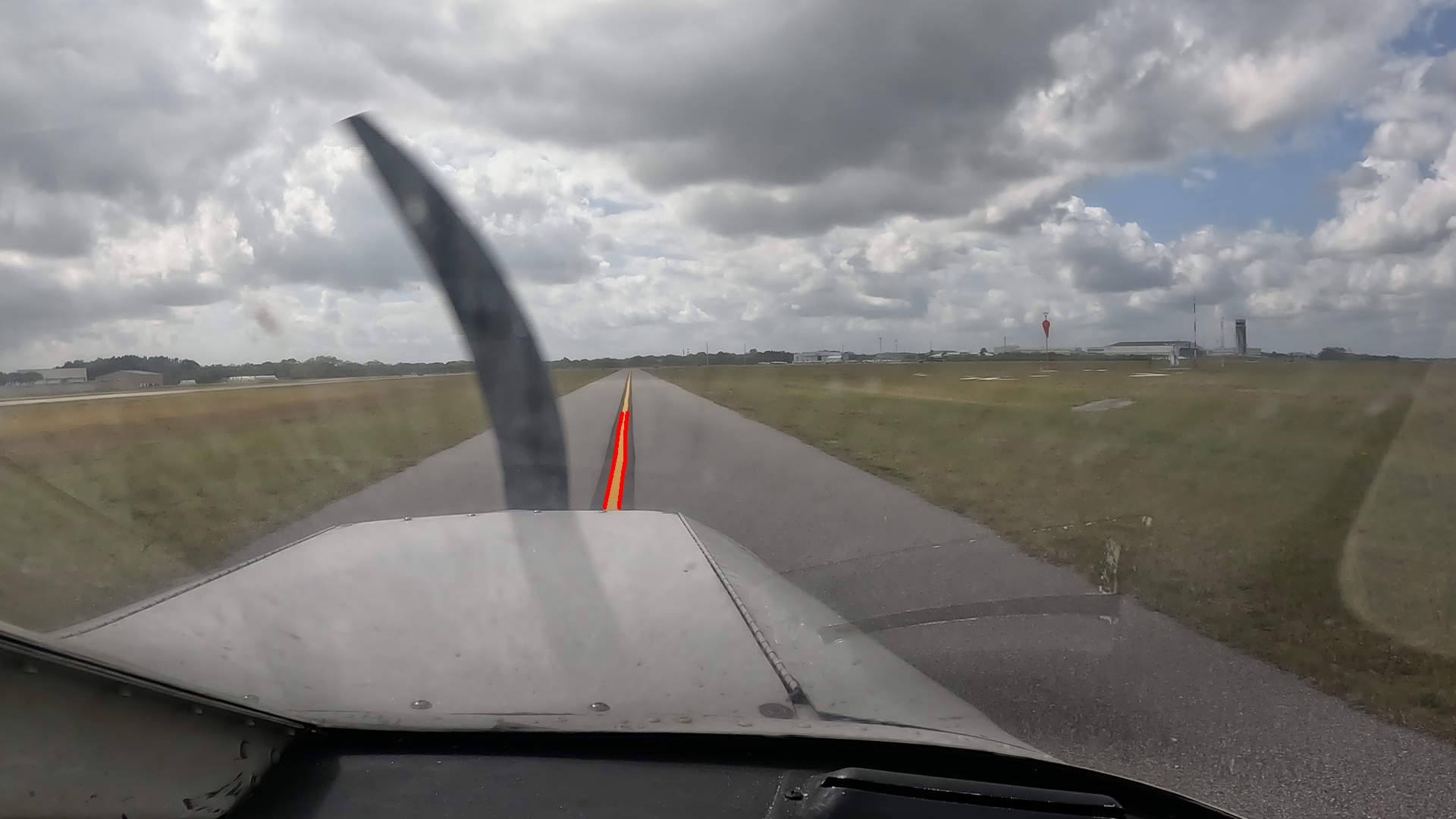}
  \caption{Straight}\label{fig:awesome_image3}
\endminipage
\minipage{0.32\textwidth}
  \includegraphics[width=\linewidth]{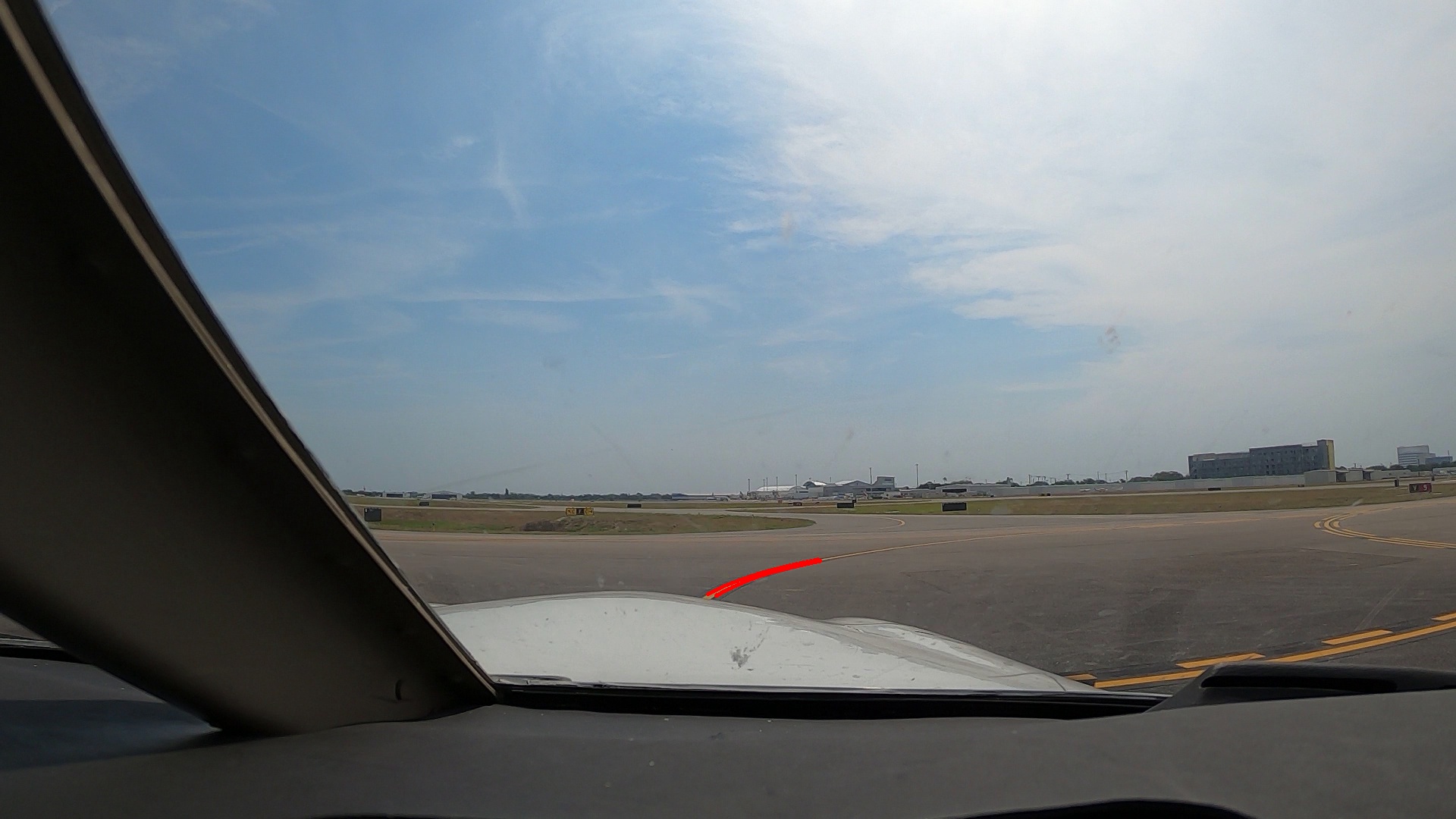}
  \caption{Right direction}\label{fig:awesome_image1}
\endminipage\hfill
\end{figure*}
The approach used for labeling images in a dataset involves a series of steps to detect and mark taxiway line markings (Algorithm \ref{alg:imgproc}). The subsequent sections outline the procedure and discusses the concepts involved.

\subsection{Drawing the Contour}
The approach for labeling images in the dataset begins with the manual drawing of a contour on the initial image (Lines 1-8). This contour precisely specifies the region where the taxiway line marking exists, acting as a reference for subsequent operations. It is important to accurately delineate the desired area to ensure precise detection of the line markings.
\begin{algorithm}[h!tbp]
\caption{Initial Approach for Labeling Images: Contour-Based Detection and Line Extraction}
\label{alg:imgproc}
\begin{algorithmic}[1]
\Require $img\_path$: Image path
\Ensure $annotated\_img$: Annotated image
\For{$img \gets img\_path$}
    \Procedure{loadImg}{$img\_path$}
          \State Read the image at $img\_path$
    \EndProcedure
    \Procedure{drawROI}{$image$, $mouse\_events$}
        \State Initialize $contour\_pts \gets []$
        \State On $mouse\_events$ (x, y) $\gets contour\_pts$ and draw ROI on img
    \EndProcedure
    
    \Procedure{applyEdgeDetection}{$image$, $mask$}
        \State Convert $img$ to grayscale
        \State Apply Bitwise AND on grayscale $img$ and $mask$
        \State Apply Gaussian blur on the img
        \State Apply Canny edge detection and return Canny img
    \EndProcedure
    
    \Procedure{applyHoughTransform}{$canny\_img$}
        \State Hough transform $\gets canny\_img$ extract $s\_lines$
    \EndProcedure
    
    \Procedure{applyCurveFitting}{$edge\_coords$, $thres$}
        \State Ramer-Douglas-Peucker ($edge\_coords$, $thres$) to extract $c\_lines \gets simpli\_coords$
    \EndProcedure
    
    \Procedure{genPixels}{$h\_lines$, $simpli\_coords$}
        \State Initialize $final\_contour \gets []$
        \State Bresenham's line ($h\_lines) \gets final\_contour$
        \State $simplified\_coords \gets final\_contour$
    \EndProcedure
    
    \Procedure{save}{$final\_contour$}
        \State Save $final\_contour$ to text file
        \State Annotate original image with $final\_contour$
    \EndProcedure
    \Procedure{output}{$annotated\_img$}
        \State Display $annotated\_img$ on screen
    \EndProcedure
\EndFor
\end{algorithmic}
\end{algorithm}
\subsection{Canny Edge Detection}
Once the contour is drawn, the attention shifts to the inside region of the contour. The Canny edge detection algorithm is applied to this region to identify the edges present (Lines 9-14). This algorithm is widely used for edge detection due to its ability to accurately identify edges while reducing noise. It involves several steps, including noise reduction, gradient calculation, non-maximum suppression, and hysteresis thresholding.
    
By applying the Canny edge detection algorithm, all the edges within the specified region are found. These edges form the basis for further analysis to detect the taxiway line markings accurately.

\subsection{Hough Transform for Line Detection}
The next step involves the use of the Hough transform (Lines 15-17). It is a popular technique in computer vision used to detect lines or shapes in an image. In the context of this approach, the Hough transform is employed to identify the line markings on the taxiway.

However, it has a limitation—it can only detect straight lines. This limitation becomes significant when dealing with curved line markings on the taxiway. To overcome this limitation, a curve fitting algorithm called the Ramer-Douglas-Peucker algorithm is used in conjunction with the Hough transform.

\subsection{Ramer-Douglas-Peucker Algorithm for Curve Fitting}
The Ramer-Douglas-Peucker algorithm is known for its ability to approximate curves by iteratively reducing the number of points required to represent the curve. It recursively identifies the most significant points on the curve, thereby reducing the number of points needed to represent the curve accurately. By incorporating this algorithm (Lines 18-20), with the Hough transform, the process can detect both straight and curved taxiway line markings.

\subsection{Pixel Generation with Ramer-Douglas-Peucker and Bresenham's Algorithm}
Using the Ramer-Douglas-Peucker algorithm, each marked pixel of the line is generated (Lines 21-25). This ensures that the curved lines are properly represented. Additionally, Bresenham's line algorithm is employed to generate all the pixels marked by the Hough transform lines. Bresenham's algorithm is an efficient method for drawing straight lines between two points, and its application here helps capture all the relevant pixels representing the line markings.

\subsection{Pooling and Saving Coordinates}
Once all the pixels representing the line markings have been extracted using the aforementioned algorithms, they are pooled together (Lines 26-32). This pooling process combines the pixels from both approaches, creating a comprehensive representation of the line markings. Figure \ref{fig:awesome_image2}, \ref{fig:awesome_image3} and \ref{fig:awesome_image1} represent different images from the dataset which have been annotated through the above process.

The resulting coordinates (x, y) of the line markings are then saved to a text file. This allows for further analysis, processing, or integration with other systems. Storing the coordinates provides a convenient way to access and utilize the labeled data for various purposes, such as training machine learning models or generating visualizations.

\subsection{Consistency through Contour Reuse}
To extend this approach to the entire dataset, the same contour drawn on the initial image is applied to other images in the dataset that capture a similar area from the same angle. By reusing the contour, the process can be automated to a certain extent, reducing the manual effort required for each image, and ensures consistency in labeling across similar images.

\subsection{Limitations and the Need for Automation}
It is important to note that the limitation of this process is the need to manually draw a new contour for every new scenario that the camera captures. It means that the approach is not entirely automated and still relies on manual intervention. Ideally, the goal would be to develop a system where specifying a region of interest for a particular camera angle would be sufficient for labeling all the images. 

Furthermore, one of the limitations with the data collection was the rapid degradation of the camera battery performance due to extended heat exposure. The current dataset lacks in diversity with images pertaining to daylight, clear sky conditions and future data collection efforts will incorporate diverse environmental and weather conditions.

\section{Conclusion}
\label{conclusion}

The AssistTaxi dataset presented in this paper is a valuable resource for advancing autonomous operations in taxiways and runways, enhancing aviation safety. It enables the development of intelligent solutions for optimizing critical airport areas. Researchers can leverage the dataset to gain insights, identify hazards, and make informed decisions for aircraft and personnel safety. The AssistTaxi dataset facilitates benchmarking algorithms, evaluating system performance, and enabling collaboration within the aviation community. We also propose an initial approach to annotating the dataset, however an automated method is required to identify the lines and annotate it. While using the dataset, users should validate the data and acknowledge its limitations in capturing all real-world scenarios. We plan to collect more data, encompassing more scenarios and test transfer learning models developed by our existing research. 

\bibliographystyle{unsrt}
\bibliography{ref}

\end{document}